\documentclass[runningheads,a4paper]{llncs}

\usepackage{amssymb}
\usepackage{graphicx}

\usepackage{url}
\urldef{\mailsa}\path|{alfred.hofmann, ursula.barth, ingrid.haas, frank.holzwarth,|
\urldef{\mailsb}\path|anna.kramer, leonie.kunz, christine.reiss, nicole.sator,|
\urldef{\mailsc}\path|erika.siebert-cole, peter.strasser, lncs}@springer.com|

\begin{document}

\mainmatter  

\title{Semantic Correspondence: A Hierarchical Approach}

\titlerunning{Akila Pemasiri}

%
%

\author{Akila Pemasiri\inst{1} \and
Kien Nguyen \inst{1} \and \\
Sridha  Sridharan\inst{1}  \and Clinton Fookes\inst{1}}
\authorrunning{Akila Pemasiri et al.}
%

\institute{Image and Video Research Lab, Queensland University of Technology, Brisbane, Australia\\
\email{\{a.thondilege,k.nguyenthanh,s.sridharan,c.fookes\}@qut.edu.au}\\
}
%

%
%

\toctitle{Lecture Notes in Computer Science}
\tocauthor{Authors' Instructions}
\maketitle

\begin{abstract}
Establishing semantic correspondence across images when the objects in the images have undergone complex deformations remains a challenging task in the field of computer vision. In this paper, we propose a hierarchical method to tackle this problem by first semantically targeting the foreground objects to localize the search space and then looking deeply into multiple levels of the feature representation to search for   point-level correspondence. In contrast to existing approaches, which typically penalize large discrepancies,  our approach allows for significant displacements, with the aim to  accommodate large deformations of  the objects in scene. Localizing the search space by semantically matching object-level correspondence, our method robustly handles large deformations of objects. Representing the target region by concatenated hypercolumn features which take into account the  hierarchical levels of the surrounding context, helps to clear the ambiguity to further improve the accuracy. By conducting multiple experiments across scenes with non-rigid objects, we validate the proposed approach, and show that it outperforms the state of the art methods for semantic correspondence establishment.

\end{abstract}

\section{Introduction}
\label{sec:intro}

In the computer vision domain, the correspondence problem refers to the process of establishing connections between similar points/regions across different images. The techniques for establishing correspondence among different images are used in many computer vision applications ranging from object tracking ~\cite{yilmaz2006object}, shape registration ~\cite{dragomir2005correlated} to  structure from motion and 3-D object reconstruction ~\cite{agarwal2009building} . This research area has evolved from establishing correspondence between spatially or temporarily adjacent images (canonical correspondence) to images from different scenes where those scenes share similar characteristics (semantic correspondence)~\cite{bristow2015dense}.

\begin{figure}[t]
\centering
\includegraphics[height=3.5cm]{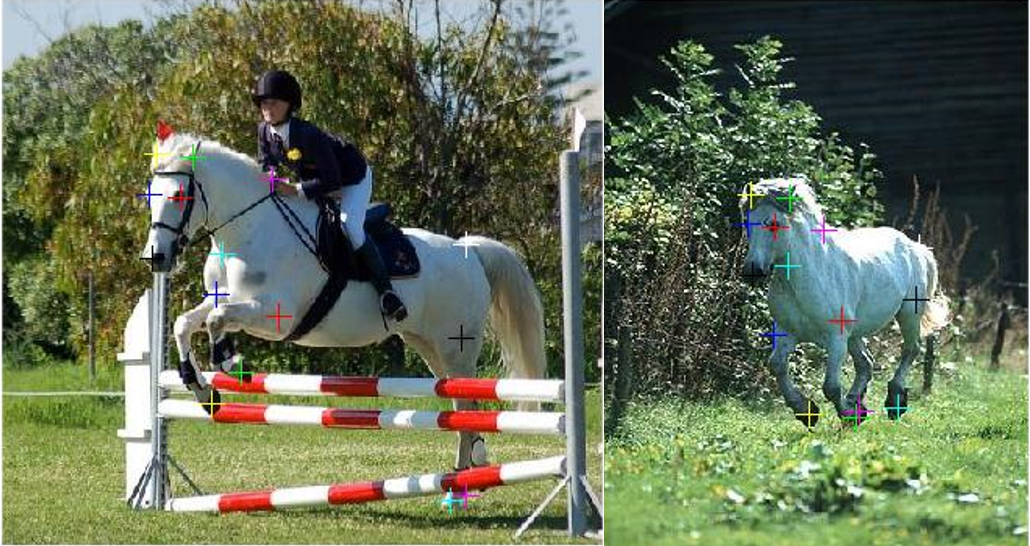}
\caption{Semantic correspondence across 2 images.}
\label{correspondenceimage2}
\end{figure}

Figure \ref{correspondenceimage2} depicts an example of semantic correspondence, where the two images contains instances from the same class) in different background settings. In semantic correspondence we aim at identifying the points in the second image  which are semantically similar to the marked points in the first image. In  addition to the changes in the background,  the objects in the environment can also go through significant intra-class appearance changes, object deformation, changes in the illumination of the scenes, shadings and different camera settings  all of which increase the complexity of the semantic correspondence problem (Figure \ref{correspondenceimage2}).

To address these challenges, we propose a novel approach where the correspondence estimation is carried out in a hierarchical manner where first the semantic segmentation is carried out to limit the search scope, while allowing the correspondence points to remain in the region corresponding to the object. Unlike the approaches proposed in the past which penalize large discrepancies in matching points ~\cite{kim2013deformable}, our approach accommodates large deformations for the objects in scene. In addition we employ a feature descriptor which encodes different levels of information about the object in the scene and the points. The descriptor consists of hypercolumn features ~\cite{hariharan2015hypercolumns}, which captures information about the objects at different levels ranging from concrete features to abstract features. In addition our descriptor includes features, extracted from Universal Correspondence Network (UCN) ~\cite{choy2016universal}, which  explicitly contains a convolutional spatial transformation layer generating a set of features that are invariant to transformations. To compare the features for correspondence, we use a 2-channel convolutional neural network ~\cite{zagoruyko2015learning} which further captures the nonlinear relations between the input and output. 

The remainder of the paper is organized as follows. In Section 2 we analyse the recent literature on scene correspondence. Section 3 describes the proposed system and its subsections elaborate each component of the system. In Section 4 we present our experimental and in Section 5 we conclude the paper with a discussion of the results on the presented method.

\section{Related Work}
From the initial work where the raw pixel intensities were used to measure (dis-)similarity and to establish the correspondence ~\cite{horn1981determining}, research on image correspondence has  advanced to obtaining robust estimation in adverse conditions such as illumination changes in the images, non-rigid deformations of the objects in the environment and degraded quality of the images ~\cite{brox2004high,mileva2007illumination}. 

Prior to the introduction of SIFT Flow, all the literature on image correspondence have used the assumption that  images under consideration share same underlying scene. With the introduction of SIFT flow, correspondence across scenes have been investigated ~\cite{liu2011sift}. SIFT Flow has evolved from the initial idea of the optical flow algorithm ~\cite{horn1981determining}. In contrast to the optical flow algorithm which takes raw pixel intensities into consideration, in SIFT Flow the SIFT descriptor ~\cite{liu2011sift} which is defined based on local gradient information is used to calculate the displacement. However being a hand-crafted feature, the SIFT descriptor does not provide optimal estimations when different shadings and illumination variations are encountered. In addition, the energy function used by the SIFT Flow enforces the flow to be as small as possible which can lead to erroneous results when there are large deformations in the objects of the image.
Deformable Spatial Pyramid (DSP) ~\cite{kim2013deformable} models the correspondence problem as a graphical model where the correspondence is measured using different levels as grid cell levels and pixel levels. First the pixel level features are calculated; then by grouping those features the first level grid cell  features are calculated and the first level grid cell  features are grouped to calculate second level. This continues up to a predefined number of levels. When establishing the correspondence, first the upper most layer correspondence is found which is then propagated to lower layers. However in the top down approach that is followed in DSP for correspondence establishment, the neighbouring features at all levels are enforced to stay in a particular spatial extent. This limits the algorithm's capacity to perform well in scenarios where the objects in the environment have gone through complex deformations.

In the work by Bristow \textit{et al.} \cite{bristow2015dense} the semantic correspondence has been established using Linear Discriminant Analysis (LDA). In their approach a linear classifier has been learnt per pixel for all the pixels  in the reference image and it has been applied on the target image using  a sliding window. In this approach classifier learning is performed on all the pixels, including foreground pixels and background pixels in the source image, and then they are applied on all the pixels in the target image.

The Universal Correspondence Network (UCN) \cite{choy2016universal} , which is currently the state of the art method for semantic correspondence establishment, uses a neural network based method for feature extraction and comparison, introducing the concept of convolutional spatial transformer which is an improved version of spatial transformer \cite{jaderberg2015spatial}. However, in this approach the all the points in the image are searched for correspondence estimation and the information from the intermediate layers of the network are not considered.

In contrast to the approaches that have been proposed to date we use a novel method to solve the correspondence estimation problem. Unlike the prior approaches which consider the whole image, we limit the alignment scope to semantic segments of the object \cite{he2017mask}. Leveraging the recent improvement in solving computer vision problems using deep networks, we use hypercolumn features \cite{hariharan2015hypercolumns} generated using Alexnet \cite{krizhevsky2012imagenet}, in combination with UCN features in the task. Due to the ability of the hypercolumn features to combine features across multiple levels, they are able to more effectively capture the relationship between pixels making it more akin to how human cognition establishes correspondence across scenes.

\section{Methodology}
\label{method}

For robust correspondence establishment we identify three questions that need to be addressed which are listed below; the existing semantic correspondence matching techniques, including ours can be categorized on how they address these 3 requirements as depicted in Table \ref{table_questions}. In this section we elaborate how our approach   tackles these questions, describing   our ``Alignment scope'' identification  in Section \ref{boundary_detection} , ``What to align'' in Section \ref{featuregeneration} and ``How to align'' in Section \ref{feature_comparison}.

\begin{enumerate}
\item Alignment scope -- which area of the image is considered as the search space in identifying the correspondence (e.g. whole image, object segment)?
\item What to align -- which features are to be aligned (e.g. pixel intensity, pixel gradients)?
\item How to align -- how to formulate the correspondence problem to obtain the parameters?
\end{enumerate}

\begin{table}[!h]
\footnotesize
\begin{center}

\begin{tabular}{|p{1.5cm}|p{2cm}|p{2.5cm}|p{2.3cm}|p{2.7cm}|}
\hline
&SIFT flow &DSP & UCN & Ours\\
\hline
\hline
Alignment Scope & Whole Image & Cells of the Image  & Whole Image & Object Region\\
What to Align & SIFT Features & Any Local Descriptors & UCN features & UCN features \& Hypercolumn Features\\
How to Align & Energy Minimization &  Multi Scale Energy Minimization & Neural Network & Neural Network\\
\hline
\end{tabular}
\end{center}
\caption{Different approaches that are used to solve the semantic correspondence estimation.}
\label{table_questions}
\end{table}

\begin{figure}
\centering
\includegraphics[height=5.5cm]{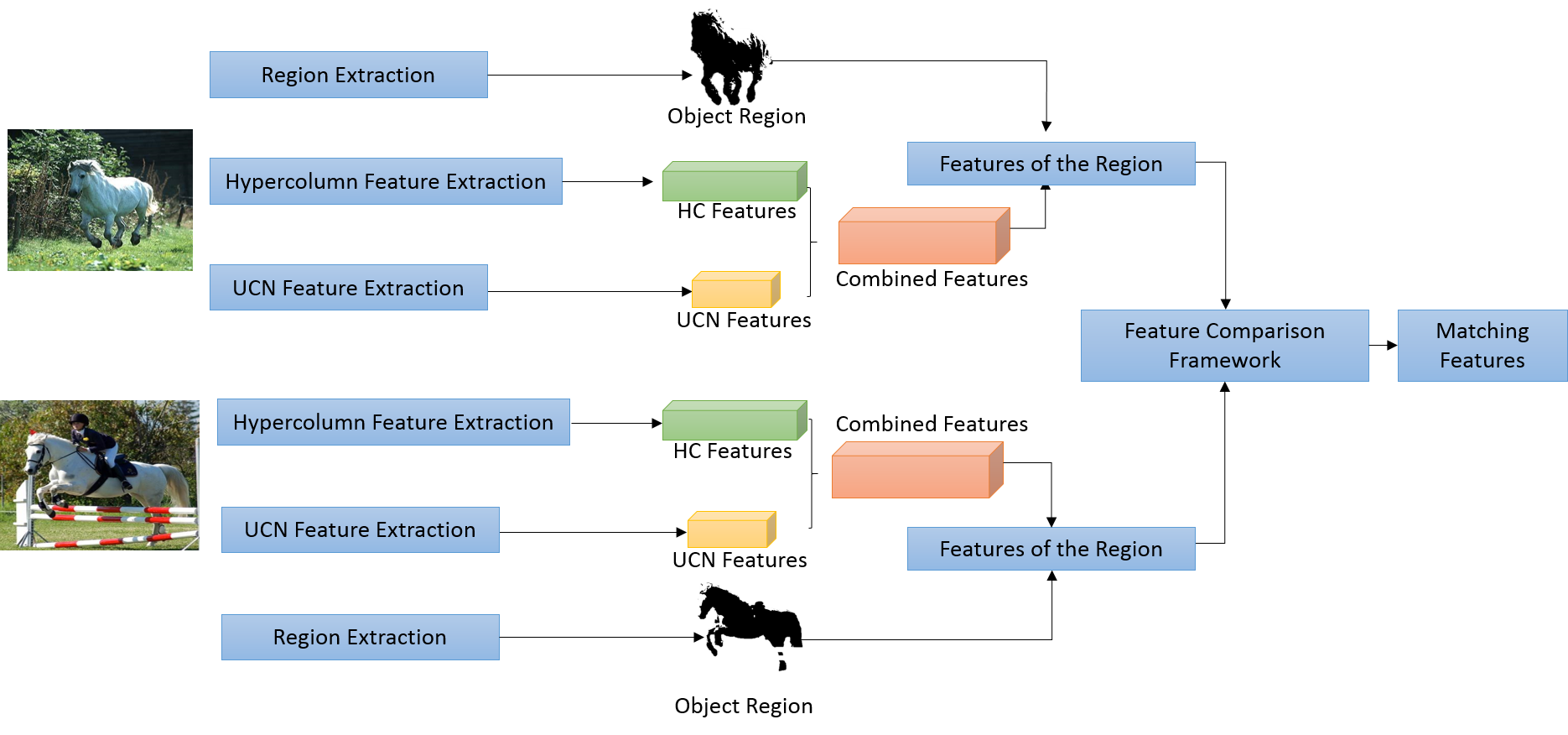}
\caption{The overall flow of the proposed system.}
\label{overall}
\end{figure}

An abstract view of the proposed framework is illustrated in Figure \ref{overall}. The object region detection and feature extraction tasks are performed concurrently, and   feature vectors of the object region are   fed into the feature comparison framework where the pairs of features are evaluated for their similarity. As can be seen, the region and the feature extraction stages are  performed at the top level of the hierarchy with the aim of limiting the search scope, while the feature comparison is  carried out in the next level of the hierarchy. 

\subsection{Semantic Object Detection}
\label{boundary_detection}
The detection of the object boundaries provide two benefits in the correspondence estimation by limiting the search scope. Firstly, when we do the point to point correspondence estimation process we force the corresponding points to lie  within the object boundary, which increases the accuracy and  secondly since we  are searching only within the boundary, this makes the estimation process faster. We used object labels and object boundaries which are detected by  Mask R-CNN \cite{he2017mask} which is an extension of Faster-RCNN \cite{ren2015faster}.  For the detection task we used the pretrained Mask R-CNN, which has been trained on COCO dataset \cite{lin2014microsoft}. Mask-RCNN returns the exact boundary of the semantic objects; however we used an extended boundaries to allow the corresponding point to be searched over the neighborhood of the the boundary.

\subsection{Feature Generation}
\label{featuregeneration}
Object region identification process described in the previous section \ref{boundary_detection} provides a coarse grain correspondence while narrowing down the search space for point to point correspondence. However to establish a finer correspondence it is required to identify the matching points in the corresponding object regions. This raises the requirement of using a feature descriptor to encapsulate the features of each point.
In this work, we have used the combination of two features such that different types of information about the object are encoded by the feature that are used for the correspondence estimation. The utilized features are the ``Hypercolumn features''(HC features) and ``UCN features''; the HC features are capable of concatenating the details about the objects in different levels and the UCN  features are capable of handling independent spatial transformations on the features.

\subsubsection{Hypercolumn Features} 
The original definition of hypercolumns is associated with neuroscience, where a column is a group of cells that respond to different aspects of visual stimulus. A hypercolumn is a collection of those cells which enables us to perceive the complete picture of a scene by accumulating different aspects \cite{hubel1962receptive}. This concept has been adopted to the computer vision field by using the output of the layers of CNNs.

In this work we used hypercolumn features as they capture both spatial information as well as the semantic information. We used the hypercolumn features extracted from the Alexnet \cite{krizhevsky2012imagenet}. Alexnet consists of 5 convolution layers having channel dimensions of 96, 256, 384, 384 and 256 respectively where different levels of abstraction are possessed by each layer. Alexnet has been trained by minimizing the cross-entropy loss function \ref{crossentropylossalex}. In defining the  loss $L$,  $y_{ic}$ is the indicator that example $i$ has label $c$, $f_{c}\left ( x_{i} \right )$ is the predicted probability of class $c$ for image $x$, $\epsilon$ is the  regularization parameter, $w$ is the weights of the network, $N$ is the number of training examples.  In creating the hypercolumn features we used these convolution layers resulting in a 1376 dimensional feature vector for each pixel. The input images are of the size of $227 \times 227$. In order to obtain same dimensional feature vector for all the pixels we used bi-linear interpolation.

\begin{equation}
L = \sum_{i=1}^{N}\sum_{c=1}^{1000}-y_{ic}logf_{c}\left ( x_{i} \right ) + \epsilon \left \| w \right \|_{2}^{2}
    \label{crossentropylossalex}
\end{equation}

\begin{figure}
\centering
\includegraphics[height=5.5cm]{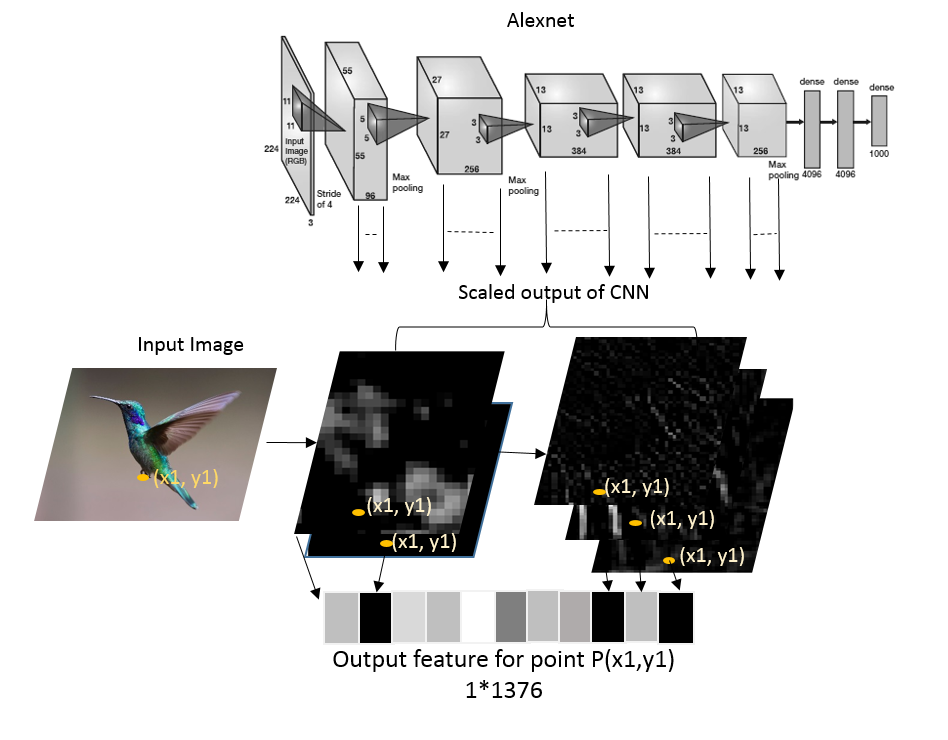}
\caption{Hypercolumn feature extraction.}
\label{hcfeatures}
\end{figure}

\subsubsection{Universal Correspondence Network  Features}

\begin{figure}[t]
\begin{tabular}{cc}
{\includegraphics[width=4.6cm]{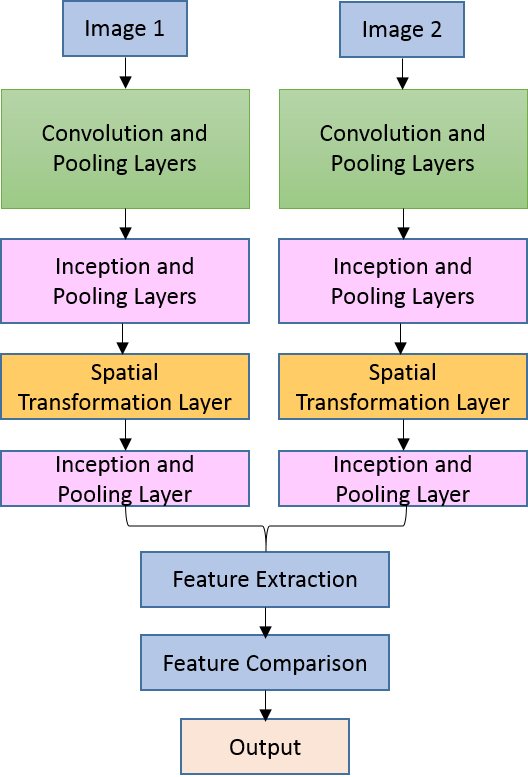}}&
{\includegraphics[width=4.6cm]{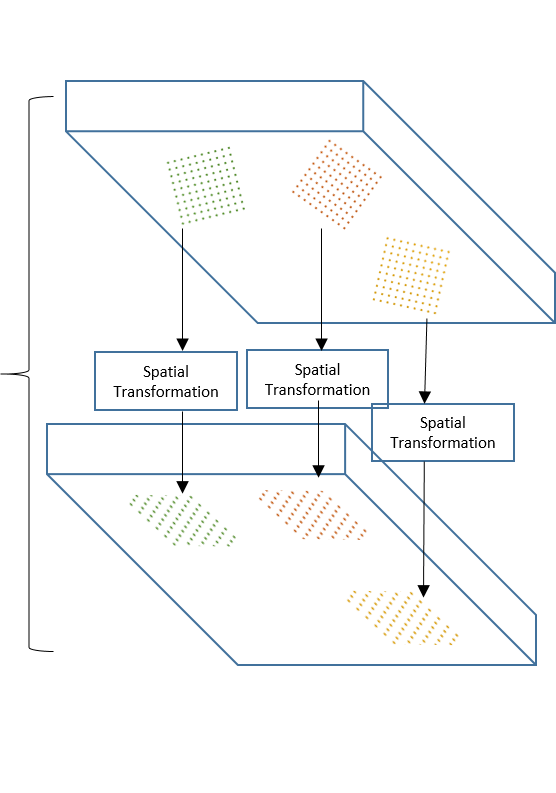}}\\
(a) UCN abstract view. &(b) Convolutional spatial transformer.
\end{tabular}
\caption{UCN architecture. }
\label{ucnarchitecture}
\end{figure}

Universal Correspondence Network (UCN) \cite{choy2016universal} currently provides the state of the art results for semantic correspondence estimation. It has an explicit spatial transformation layer that handles not only the global transformations, but also the independent transformations of keypoints which is vital for estimating correspondence in the presence of  large intra-class deformations. We used the output features of UCN as a part of the feature descriptor in our approach to leverage this favourable characteristics of UCN. The abstract view of the Universal Correspondence network is depicted in Figure \ref{ucnarchitecture}, where the primary layers of the network are illustrated in Figure \ref{ucnarchitecture}\{a\} and the  convolutional spacial transformer that applies an independent transformation to the features is illustrated in \ref{ucnarchitecture}\{b\}.  In training the network they have used correspondence contrastive loss (Equation \ref{ucn_train}) function, where $F_{I}\left ( x_{i} \right )$ denotes the feature in image $I$ at location $x$  and $s$ = $1$ if the points under consideration are corresponding to each other. In this equation the loss is denoted by $E$, the number of training examples is denoted by $N$ and $m$ denotes the  margin that is used for the hard negative mining in the training process.

\begin{equation}
E = \frac{1}{2N}\sum_{i}^{N} s_{i} \left \| F_{I}\left ( x_{i} \right ) - F_{I^{'}} \left ( x_{i}^{'} \right )   \right \|^{2} +\left(1-s_{i} \right ) max \left(0, m- \left \| F_{I}\left ( x_{i} \right ) - F_{I^{'}} \left ( x_{i}^{'} \right )   \right \|^{2} \right)
    \label{ucn_train}
\end{equation}

\subsection{Point to Point Image Correspondence}
\label{feature_comparison}
The features obtained through UCN feature extraction and Hypercolumn feature extraction are used to train a neural network in the supervised manner by minimizing the binary cross entropy error (Equation \ref{binarycrossentropy}) using Adam optimizer \cite{kingma2014adam} which is an improved version of Stochastic gradient decent. In Equation \ref{binarycrossentropy}, $y_{i}$ which is the label for each point pair can have the values 1/0, and $\hat{y_{i}} $ stands for the output of the neural network where $N$ is the number of point features that was subjected to training.

\begin{equation}
 E = -\frac{1}{N} \sum_{n=1}^{N} \left [ y_{n}\,\,log \,\, \hat{y_{n}} + (1 - y_{n})\,\, log \,\, (1 - \hat{y_{n}}) \right ]
    \label{binarycrossentropy}
\end{equation}

We used a 2-channel architecture where the network is fed with feature patches around each pixel. The use of neural network in the point correspondence enables us to model the nonlinear relations between the input and output with the used of hidden neurons. In the 2-channel architecture, the two features which are subjected to the comparison are fed into the network as a single image with 2-channels. The efficacy of 2-channel architecture which has been demonstrated in patch feature comparison \cite{zagoruyko2015learning} motivated us to use it in semantic feature comparison work. Note that, Recurrent Neural Network(RNN) and Long Short Term Memory(LSTM) based architectures   were not considered as they are more suited to  sequence predictions \cite{le2015tutorial} where as in semantic correspondence estimation process, the sequential  information that these networks are able to capture is not important.

\section{Evaluations}
\subsection{Datasets}
For this study we used PASCAL dataset \cite{pascal-voc-2011} with Berkley key point annotation \cite{berkley}, which contains images with anatomically similar objects  and their point to point groundtruth  key point annotations. In addition, we used the Caltech-UCSD Birds 200 (CUB-200) \cite{WelinderEtal2010} database.

\subsection{Training the Comparison Network}

\begin{figure}
\begin{tabular}{cc}
{\includegraphics[width=5.99cm]{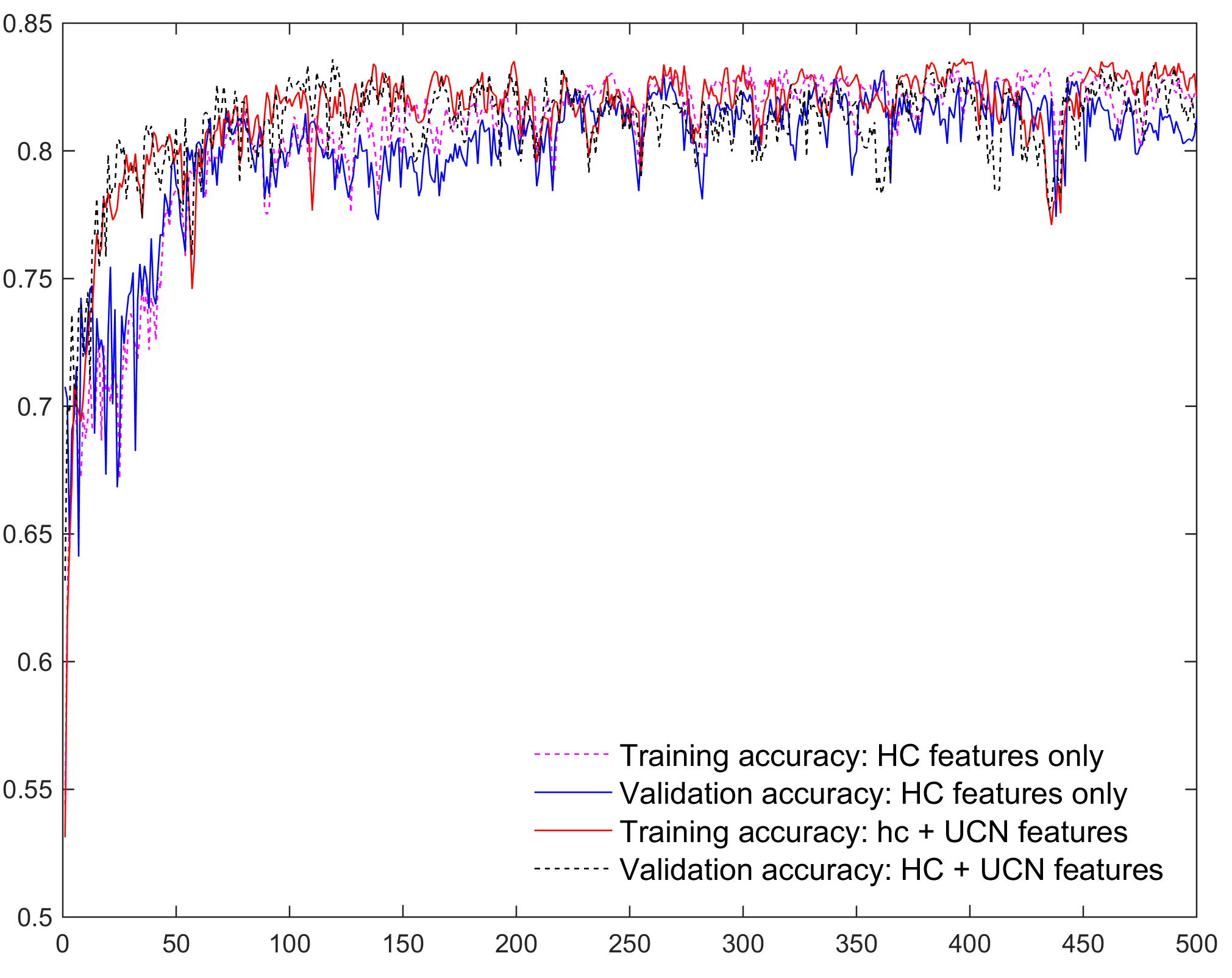}}&
{\includegraphics[width=5.99cm]{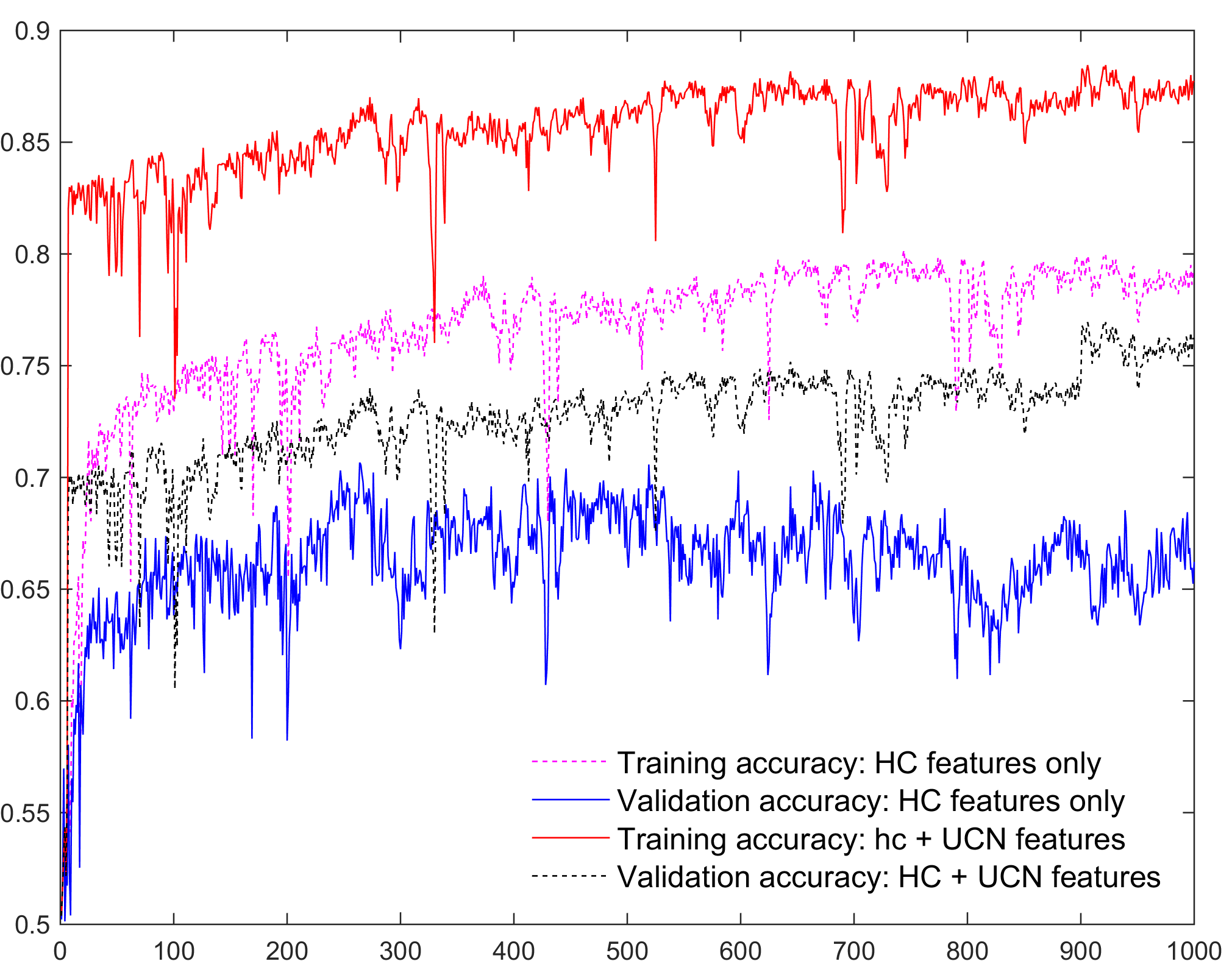}}\\
(a) For Caltech-UCSD Birds 200 dataset. &(b) For PASCAL dataset.
\end{tabular}
\caption{Accuracy plots. }
\label{accuracy_plots}
\end{figure}

The comparison network was trained on both of the dataset, with an equal number of positive and negative examples. We evaluated the accuracies for two scenarios: (i). where only the hypercolumn features are fed alone into the neural network and (ii). where the UCN features and  Hypercolumn features  are concatenated and fed into the network. Training and validation accuracies for these scenarios for the Caltech-UCSD dataset are depicted in Figure \ref{accuracy_plots}\{a\} and for the Pascal-Berkeley dataset are depicted in Figure \ref{accuracy_plots}\{b\}. In Figure \ref{accuracy_plots}\{b\} the depicted training and validation accuracies are the average of the values that were obtained for each of 20 classes in the Pascal-Berkeley dataset.

\subsection{Evaluation Metrics}
To benchmark our method with the state of the art methods, we used Percentage of Correct Keypoints (PCK) metric \cite{long2014convnets,zhou2015flowweb} which is also denoted as ``accuracy@T'' \cite{revaud2016deepmatching}. For a query image, we extract the  features of the keypoints,and  feed them into our comparison framework along with the target image features. Based on the probabilities that are assigned by the neural network, we assign the point with highest probability as the corresponding point. In the evaluations, if the identified keypoint is within the range of T from the groundtruth point, we classified it as a correct keypoint. The range has been defined as a percentage of the image dimension, where the percentage is denoted as $\alpha$. 

We have compared our method with UCN \cite{choy2016universal}, which is the state of the art method for semantic correspondence with its different settings where ``UCN HN'' refers to the UCN method with their hard negative mining technique and ``UCN HN ST'' refers to the UCN method with the hard negative mining and spatial transformation. In addition, we have compared our method with the other methods all of which have been previously compared with the UCN method for benchmarking, including the method by Long et al.  \cite{long2014convnets}, SIFT flow \cite{liu2011sift}, DSP \cite{kim2013deformable} and WarpNet \cite{kanazawa2016warpnet}.  

\begin{figure}[!h]
\centering
\includegraphics[width=9.5cm]{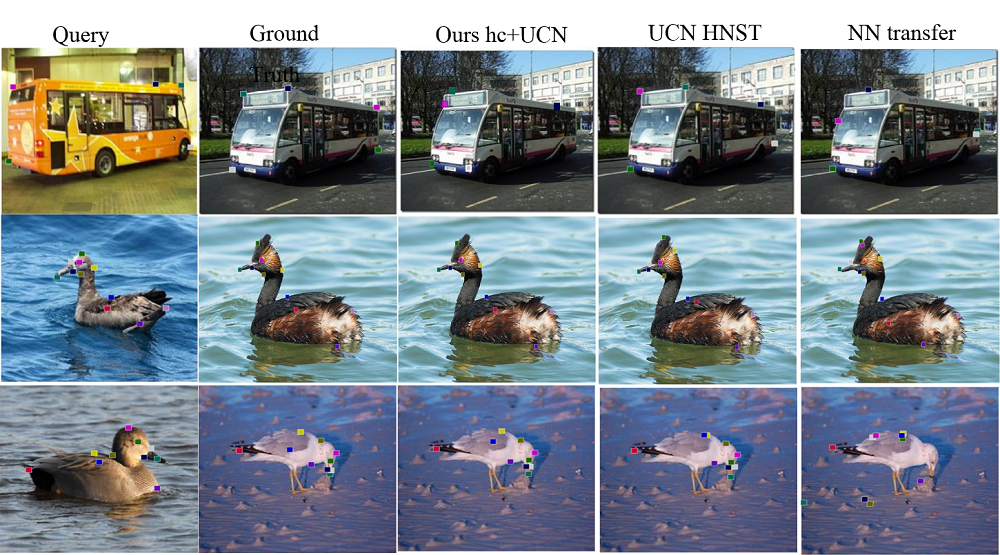}
\caption{Examples for qualitative results for semantic correspondence.}
\label{qualitative}
\end{figure}

In Table \ref{table_pck} it can be seen that for most of the cases, as well as  overall  in Pascal-Berkeley dataset our method has outperformed the state of the art. When comparing our results on Caltech-UCSD dataset, it can be seen that our method has out performed the state of the art for all the considered $\alpha$ values. Furthermore, to provide an overview of the qualitative results, Figure \ref{qualitative} depicts the corresponding keypoints that have been obtained by different algorithms including ours. 

\begin{table}
\begin{center}
\resizebox{\textwidth}{!}{%
\begin{tabular}{|l|c|c|c|c|c|c|c|c|c|c|c|c|c|c|c|c|c|c|c|c|c|}
\hline
&aero&bike&bird&boat&bottle&bus&car&cat&chair&cow&table&dog&horse&mbike&person&plant&sheep&sofa&train&tv&mean\\
\hline
\hline
conv4 flow&28.20&\textbf{34.10}&20.40&17.10&50.60&36.70&20.90&19.60&15.70&25.40&12.70&18.70&25.90&23.10&21.40&40.20&21.10&14.50&18.30&33.30&24.9\\
SIFT flow&27.60&30.80&19.90&17.50&49.40&36.40&20.70&16.00&16.10&25.00&16.10&16.30&27.70&\textbf{28.30}&20.20&36.40&20.50&17.20&19.90&32.90&24.7\\
NN transfer&18.30&24.80&14.50&15.40&48.10&27.60&16.00&11.10&12.00&16.80&15.70&12.70&20.20&18.50&18.70&33.40&14.00&15.50&14.60&30.00&19.9\\
UCN RN&31.50&19.60&30.10&23.00&53.50&36.70&34.00&33.70&22.20&28.10&12.80&33.90&29.90&23.40&38.40&39.80&38.60&17.60&28.40&60.20&31.8\\
UCN HN&36.00&26.50&31.90&31.30&56.40&38.20&36.20&34.00&25.50&31.70&18.10&35.70&32.10&24.80&41.40&46.00&45.30&15.40&28.20&65.30&35.0\\
UCN HN-ST&37.70&30.10&42.00&31.70&62.60&35.40&38.00&41.70&27.50&34.00&17.30&41.90&38.00&24.40&47.10&52.50&47.50&18.50&40.20&70.50&38.9\\
Ours HC only&42.30&31.80&43.12&32.60&64.56&39.00&39.30&43.30&28.80&35.30&18.60&42.13&38.56&26.21&48.21&52.21&48.13&18.21&41.39&70.56&40.2\\
Ours HC and UCN&\textbf{43.7}&32.3&\textbf{45.3}&\textbf{34.21}&\textbf{66.68}&\textbf{39.7}&\textbf{41.6}&\textbf{44.6}&\textbf{31.67}&\textbf{37.2}&\textbf{18.87}&\textbf{43.24}&\textbf{38.90}&27.53&\textbf{48.59}&\textbf{53.37}&\textbf{49.87}&\textbf{19.05}&\textbf{42.76}&\textbf{72.31}&\textbf{41.6}\\

\hline
\end{tabular}}
\end{center}
\caption{PCK on Pascal-Berkeley dataset for semantic correspondence with $\alpha$ = $0.1$, $L$ = $max(w,h) $.}
\label{table_pck}
\end{table}

\begin{figure}[h]
\centering
\includegraphics[width=7.5cm]{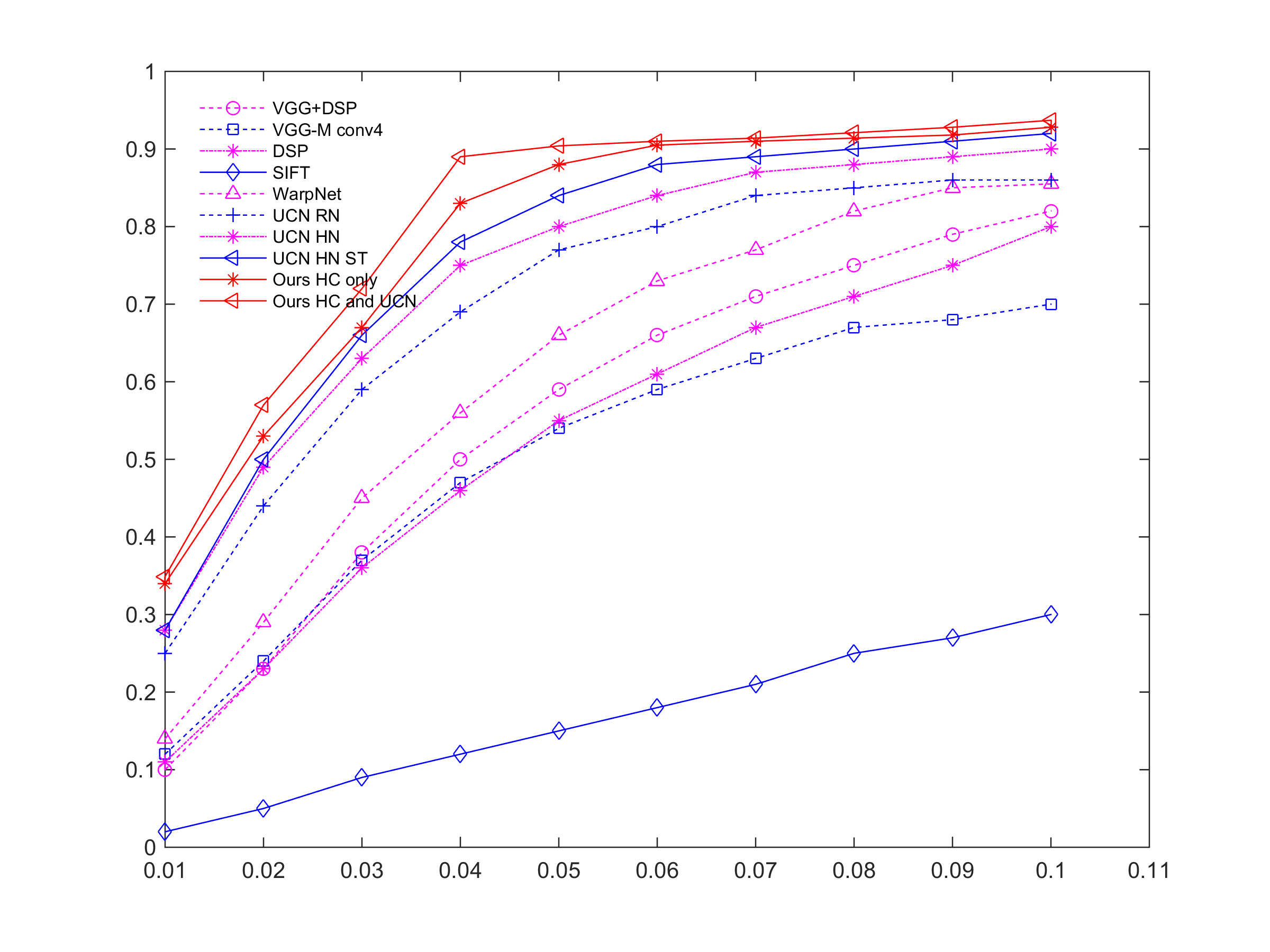}
\caption{PCK on  Caltech-UCSD dataset.}
\label{pck}
\end{figure}

When considering the qualitative results, we noticed that when the features are sharp (eg: edge or corner features) and orientation of the objects in the image have changed significantly, the performance of our method deteriorates. An example of such situation is depicted in Figure \ref{qualitative}  first row, where the image orientation of the bus in the scene has changed. However, when compared with UCN results  our method performs better in the cases where the image texture contains many variations, and one instance of such scenario is depicted in the third raw Figure .
\ref{qualitative}.
\section{Conclusion}
We have proposed a novel method for  semantic correspondence estimation, where a new feature  concatenates spatial details on the image ranging from very concrete levels to abstract levels. The hierarchical framework that we have proposed, first limits the search scope for correspondence by identifying the object boundaries. Searching within the object boundary enforces the points to lie within the object while restricting the number of comparisons. In addition, utilizing a feature that can encode the spatial transformation of each individual point on the image has yielded  improved accuracy. We have utilized a deep neural network with 2-channel architecture for feature comparison, as it provides more flexibility compared to its siamese counterpart by processing two features jointly. We have demonstrated that our approach outperforms the state of the art approaches for semantic correspondence by experiments conducted on commonly used databases. In future work, we will investigate how this novel feature combination and the comparison method can be used in the semantic correspondence estimation among the multimodal images (eg: X-ray and visual).

\end{document}